\documentclass[10pt,twocolumn,letterpaper]{article}

\usepackage[pagenumbers]{cvpr} 

\usepackage[pagebackref,breaklinks,colorlinks,citecolor=cvprblue]{hyperref}

\usepackage{multirow}
\usepackage{bbding}
\usepackage{pifont}
\usepackage{wasysym}
\usepackage{amssymb}
\usepackage{bm}

\usepackage{nicematrix}

\usepackage[utf8]{inputenc} %
\usepackage[T1]{fontenc}    %
\usepackage{url}            %
\usepackage{booktabs}       %
\usepackage{amsfonts}       %
\usepackage{nicefrac}       %
\usepackage{microtype}      %

\usepackage{fontawesome5} %
\usepackage{multirow} 
\usepackage{amsmath} %
\usepackage{xspace} %
\usepackage{titletoc} %
\usepackage{colortbl}

\definecolor{cvprblue}{rgb}{0.21,0.49,0.74}

\newcommand{\ourM}{CoVR-BLIP\xspace}    %

\definecolor{badcolor}{rgb}{1.0, 0.94, 0.97}
\definecolor{goodcolor}{rgb}{0.94, 1.0, 0.97}
\definecolor{ourcolor}{rgb}{0.807, 0.725, 0.847}
\definecolor{bluecolor}{RGB}{192, 237, 254}
\definecolor{yellowcolor}{RGB}{255, 239, 203}
\definecolor{backcolour}{rgb}{0.95,0.95,0.92}
\definecolor{lavanda}{RGB}{144,101,202}

\definecolor{ourcolor}{rgb}{0.807, 0.725, 0.847}

\def\paperID{8744} 
\def\confName{CVPR}
\def\confYear{2024}

\title{Composed Video Retrieval via Enriched Context and Discriminative Embeddings}

\author{
\textbf{Omkar Thawakar$^{1}$} \quad
\textbf{Muzammal Naseer$^{1}$} \quad
\textbf{Rao Muhammad Anwer$^{1,2}$} \quad
\textbf{Salman Khan$^{1,3}$} \quad \\
\textbf{Michael Felsberg$^{4}$} \quad
\textbf{Mubarak Shah$^{5}$} \quad
\textbf{Fahad Shahbaz Khan$^{1,4}$} \vspace{0.2em} \\
$^{1}$Mohamed bin Zayed University of AI \quad $^{2}$Aalto University \quad $^{3}$Australian National University \\
$^{4}$Link\"{o}ping University \quad $^{5}$University of Central Florida
}

\begin{document}
\maketitle

\begin{abstract}

Composed video retrieval (CoVR) is a challenging problem in computer vision which has recently highlighted the integration of modification text with visual queries for more sophisticated video search in large databases. Existing works predominantly rely on visual queries combined with modification text to distinguish relevant videos. However, such a strategy struggles to fully preserve the rich query-specific context in retrieved target videos and only represents the target video using visual embedding. We introduce a novel CoVR framework that leverages detailed language descriptions to explicitly encode query-specific contextual information and learns discriminative embeddings of vision only, text only and vision-text for better alignment to accurately retrieve matched target videos. Our proposed framework can be flexibly employed for both composed video (CoVR) and image (CoIR) retrieval tasks. Experiments on three datasets show that our approach obtains state-of-the-art performance for both CovR and zero-shot CoIR tasks, achieving gains as high as around 7\% in terms of recall@K=1 score. Our code, models, detailed language descriptions for WebViD-CoVR dataset are available at \url{https://github.com/OmkarThawakar/composed-video-retrieval}.

\end{abstract}

\section{Introduction}
\label{sec:intro}

Composed image retrieval (CoIR) is the task of retrieving matching images, given a query composed of an image along with natural language description (text).
Compared to the classical problem of content-based image retrieval that utilizes a single (visual) modality, composed image retrieval (CoIR) uses multi-modal information (query comprising image and text) that aids in alleviating miss-interpretations by incorporating user's intent specified in the form of language descriptions (e.g., text-based modification to the query image). Following CoIR, composed video retrieval (CoVR) has been recently explored in the literature \cite{ventura2023covr} where the multi-modal search is performed to retrieve \textit{videos} that display almost identical visual characteristics with the desired user intent, given a query image of a specific visual theme along with the modifier (change) text. CoVR is a challenging problem with various real-world applications, e.g.,  e-commerce and fashion, internet video search, finding live events in specific locations, and retrieving sports videos of particular players. In this work, we investigate the problem of composed video retrieval (CoVR).

The problem of CoVR poses two unique challenges: a) bridging the domain gap between the input query and the modification text, and b) simultaneously aligning the multi-modal feature embedding with the feature embedding of the target videos that are inherently dynamic. Further, their context can also vary across different video frames. To address the problem of CoVR, the recent work \cite{ventura2023covr} introduces an annotation pipeline to generate video-text-video triplets from existing video-caption datasets. The curated triplets contain the source and target video along with the change text describing the differences between the two videos. These triplets are then used to train a CoVR model, where a multi-modal encoder encodes the image query to obtain visual features which are passed along with the change text to an image-grounded text encoder, thereby generating the feature embedding. In this way, a correspondence is established between the latent embedding of input visual query's and the desired change text to retrieve a target video.

\begin{figure*}[!t]
  \centering\small
   \includegraphics[width=\textwidth]{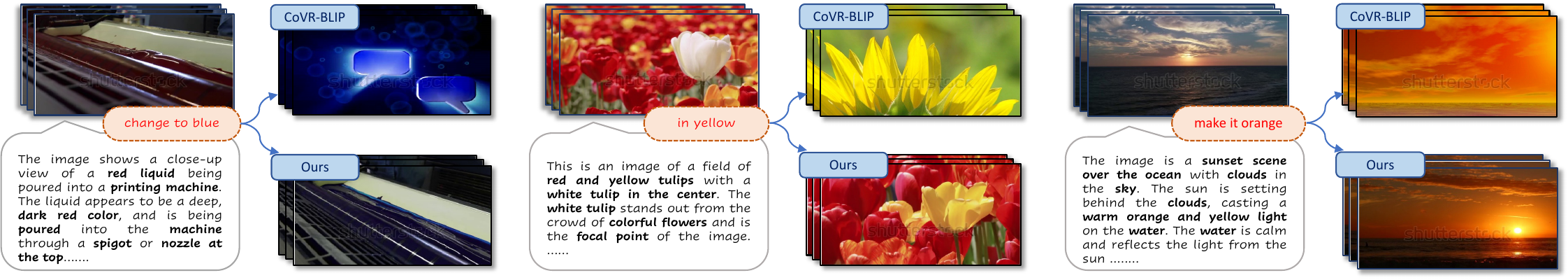}
   \caption{Comparison between the baseline CoVR-BLIP \cite{ventura2023covr} (top row) and our approach (bottom row) on example video samples from the WebVid-CoVR testset. Here, the change text is highlighted in red. We observe that the baseline typically focuses only on the change while ignoring the semantic alignment of the target video with the query input video (e.g., the composed target video in the second example from the left should have change "yellow" reflected on the salient white tulip surrounded by red flowers, as in the query input). However, the retrieved target video loses the context (red flowers surrounding the yellow tulip). This suggests that it is particularly challenging for the model to understand the correspondence between the change text and the relevant target video using \textit{only} the visual input. In contrast, our retrieved target videos are visually similar to the input query composed with the change text. Our approach leveraging detailed descriptions (highlighted in white boxes) for joint multi-modal embedding alignment encodes the necessary context to alter the composition of the video (e.g., changing the color of the "white flower" in 2nd video to yellow and changing the color of the "sky and clouds" to orange in 3rd video).
   }
   \label{fig:language descriptions demo}
\end{figure*}

We note that the aforementioned framework \cite{ventura2023covr} struggles (see \cref{fig:language descriptions demo}) since 
the latent embedding of a query visual input (image/video) is likely to be insufficient to provide necessary semantic details about the query image/video due to the following reasons: a) visual inputs are high-dimensional and offer details, most of which are not related to the given context, b) the visual depiction often shows a part of the broader context and there exist non-visual contextual cues that play a crucial role in understanding the given inputs. This motivates us to look into an alternative approach that explicitly encodes contextual information beyond what is apparent through only the visual input. 

In this work, we argue that the detailed language descriptions of the visual content is likely to provide complementary contextual information that is otherwise  difficult to encode through visual input only. For instance, consider example query videos of red liquid, flowers, and sunset in \cref{fig:language descriptions demo}. We can observe that the context becomes clear with language descriptions of these query videos; the red liquid is for the printing machine not immediately visible in the input, the white tulip stands out from the crowd of colorful red and yellow flowers, and the sun is setting over the ocean and behind the clouds. Here, richer semantics and a better context reduce the ambiguities while emphasizing important relationships e.g., the saliency of the white tulip means the change text relates to its color change, existing colors of the sky at sunset mean the change to orange should pertain to it. 

\noindent \textbf{Contributions:} We propose a framework that explicitly leverages detailed language descriptions to preserve the query-specific contextual information, thereby reducing the domain gap with respect to the change text for CoVR. To this end, we utilize recent multi-modal conversational model to generate detailed textual descriptions which are then used during the training to complement the query videos. Furthermore, we learn discriminative embeddings of vision, text and vision-text during contrastive training to align the composed input query and change text with target semantics for enhanced CoVR. Our framework can be flexibly employed for both CoVR and CoIR tasks. 

Extensive experiments on three datasets reveal the merits of our proposed contributions leading to state-of-the-art performance on both CoVR and zero-shot CoIR tasks. On the WebVid-CoVR dataset, our approach achieves a significant gain of $\approx$7\% in terms of recall@K=1 score compared to the recent CoVR-BLIP \cite{ventura2023covr}. On the CIRR test set for the zero-shot setup, our approach achieves recall@K=1 score of 40.12. \cref{fig:language descriptions demo} shows a comparison on example WebVid-CoVR test set example video samples between our approach and the recent CoVR-BLIP method \cite{ventura2023covr}.

\section{Related Work}

\noindent \textbf{Composed Image Retrieval (CoIR): } A significant progress has been made in the field of content-based image retrieval thanks to recent advances in deep learning techniques~\cite{chopra2005learning,wang2014learning,gordo2016deep,radenovic2016cnn}.
The problem holds extensive practical significance finding applications in diverse domains such as, product search, face recognition, and image geo-localization ~\cite{liu2016deepfashion,schroff2015facenet,parkhi2015deep,hays2008im2gps}. Following the advances in cross-modal image retrieval, the scope has been extended to multiple query modalities such as, text-to-image retrieval, sketch-to-image retrieval, cross-view image retrieval, event detection and also to the problem of composed image retrieval (CoIR)~\cite{wang2016learning,sangkloy2016sketchy,lin2015learning,jiang2015bridging,vo2019composing}. CoIR is challenging since it requires image retrieval based on its reference image and corresponding relative change text. Most existing CoIR approaches are built on top of CLIP~\cite{radford2021learning} and learn the multi-model embeddings comprising reference image and relative change text caption for target image retrieval ~\cite{CLIP2TV,CLIP2Video,BridgeFormer,luo2022clip4clip,liu2022ts2}. These methods carefully harness the capabilities of large-scale pretrained image and text encoders, effectively amalgamating compositional image and text features to achieve improved performance. \\
\noindent \textbf{Composed Video Retrieval (CoVR): } The field of text-to-video retrieval has witnessed significant breakthroughs as a pivotal sub-domain within the broader context of multimedia information retrieval~\cite{CLIP2Video,CLIP2TV,BridgeFormer,liu2022ts2,luo2022clip4clip,ma2022x}. Early efforts in this domain predominantly explored content-based retrieval approaches, leveraging key-frame analysis, color histograms, and local feature matching. The advent of deep learning techniques has further revolutionized text-to-video retrieval, with the emergence of multi-modal embeddings and attention mechanisms ~\cite{rasheed2023fine,xu2021videoclip,xue2022clip,yang2021taco,yao2021filip}. Recently, \cite{ventura2023covr} explored the problem of composed video retrieval (CoVR) where the objective is to retrieve the target video, given the reference video and its corresponding compositional change text. Due to the unavailability of a benchmark and following existing CoIR works \cite{brooks2022instructpix2pix,cirr}, \cite{ventura2023covr} propose a new benchmark for CoVR, named WebVid-CoVR, which comprises a synthetic training set and a manually curated test set. Further, the authors also propose a framework, named CoVR-BLIP, that is built on top of BLIP~\cite{li2023blip} where an image grounded text encoder is utilized to generate multi-model features and aligns it with target video embeddings using a contrastive loss \cite{hn_nce}.   \\
\textbf{Our Approach:} Different from COVR-BLIP \cite{ventura2023covr}, our approach leverages detailed language descriptions of the reference video that are automatically generated through a multi-modal conversation model and provide with following advantages. First, it helps in preserving the query-specific contextual information and aids in reducing the domain gap with the change text. Second, rather than relying on only using the visual embedding to represent target video as in \cite{ventura2023covr}, learning discriminative embeddings through vision, text, and vision-text enables improved alignment due to the extracting complementary target video representations. It is worth mentioning that these automatically generated detailed language descriptions can be effectively utilized within our framework either only during training or at both training and inference. In both cases, our approach leads to superior performance compared to original \cite{ventura2023covr} as well as using default (short) text captions within \cite{ventura2023covr}. Furthermore, our approach exhibits notable competitive capabilities in both transfer learning and zero-shot learning contexts for the CoIR task.

\section{Method}
\label{sec:Method}
\begin{figure*}[!t]
  \centering
   \includegraphics[width=\textwidth]{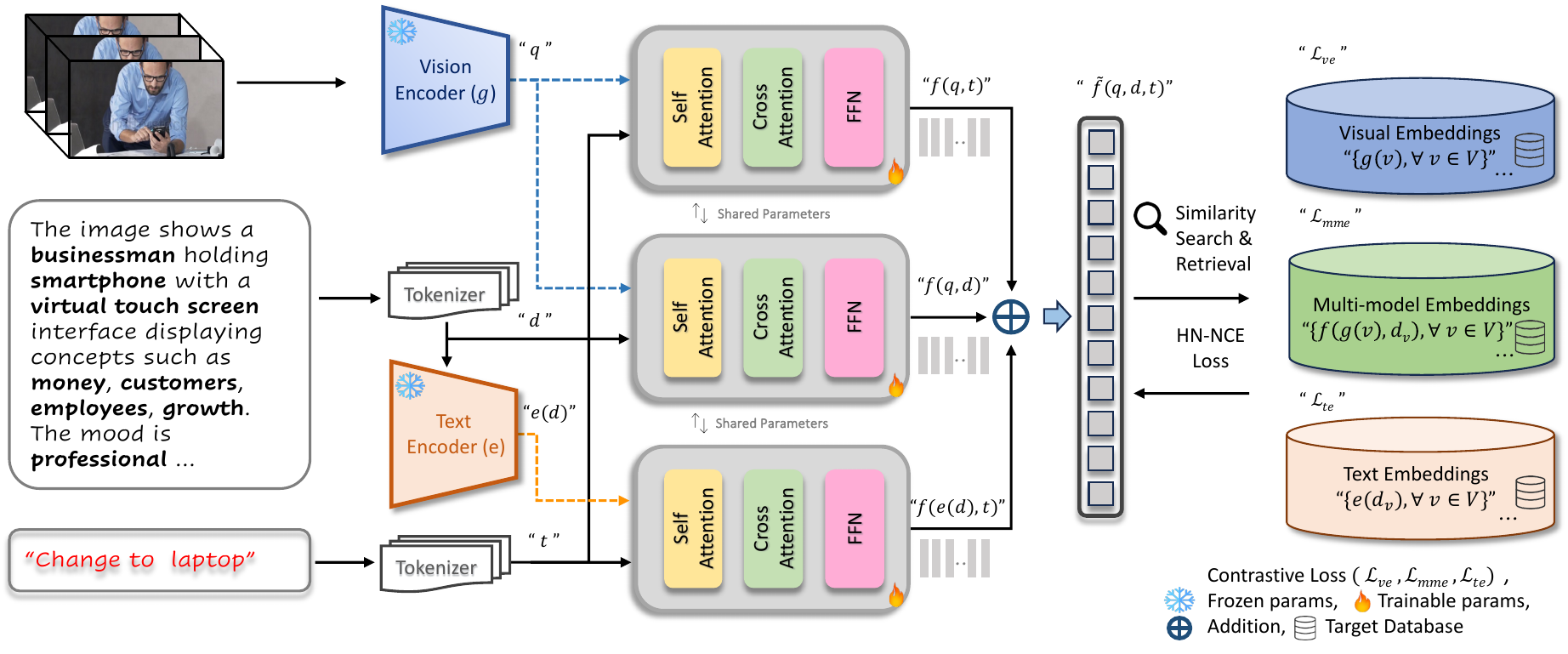}
   \caption{
   Our framework comprises three inputs: the reference video, a detailed visual description of an input video, and a change text corresponding to the target video. The input video is encoded by the vision encoder $g$, and the description is encoded by the frozen text encoder $e$. The default tokenizer tokenizes the change text. The encoded input triplet ($q$,$d$,$t$) is then processed by the multi-model encoder ($f$) grounding two inputs a time. The dotted lines shown are going to the cross-attention for grounding. During training, we add outputs of the multi-model encoder to obtain the joint multi-model embedding $\Tilde f(q,d,t)$ that is aligned across three target databases using hard negative contrastive losses (HN-NCE): $\mathcal{L}_{ve}$, $\mathcal{L}_{mme}$, and $\mathcal{L}_{te}$. During inference, our approach can utilize input query or a combination of input query along with its description to retrieve a composed target video.}
   \label{fig:architecture}
\end{figure*}

\textbf{Problem Statement:} Composed Video Retrieval (CoVR) strives to retrieve a target video from a database. This target video is desired to be aligned with the visual cues from a query video but with the characteristics of the desired change represented by the text. Formally, for a given embedding of input query $q \in Q$ and the desired modification text $t\in T$, we optimize for a multi-modal encoder $f$ and a visual encoder $g$, such that $f(q,t)\approx g(v)$, where $v\in V$ is the target video from a database. As discussed earlier, the problem of CoVR is challenging since it requires bridging the domain gap between input query $q$ and the modification text $t$. Furthermore, it requires simultaneously aligning the multi-modal feature embedding $f(q,t)$ with the feature embedding of target videos that are inherently dynamic, and their context also varies across different video frames. \\
\textbf{Baseline Framework:} To address the above problem, we base our method on the recently introduced framework \cite{ventura2023covr}, named CoVR-BLIP, that trains the multi-modal encoder $f$ which takes the representations from the visual encoder $g$. The visual encoder $g$ remains frozen and is used to get the latent embeddings for visual input query which are then provided to multi-modal encoder $f$ along with the tokenized change text $t$ to produce multi-modal embedding $f(q,t)$. Then, the input visual query and the change text $t$ are aligned with the desired target videos using a contrastive loss between $f(q,t)$ and $g(v)$ (\cref{fig:architecture}). This results in a direct correspondence between the visual latent embedding of the input query and the desired change text for retrieving a target video. For more details, we refer to \cite{ventura2023covr}.

We note that the baseline CoVR-BLIP framework struggles to effectively preserve the contextual information of the query sample, since the multi-modal information is likely biased towards the change text. This is evident in \cref{fig:language descriptions demo}, where the dominant feature in the baseline is the change text mixed with the holistic representation of the visual query e.g., yellow follower or orange sky. Instead, here the objective was to convert only the white tulip to yellow with the surrounding red flowers or orange sky over the beach. Next, we propose our approach that aims to alleviate these limitations for improved CoVR performance.

\subsection{Architecture Design}
\textbf{Motivation: } To motivate our proposed approach, we distinguish two desirable characteristics that are to be considered when designing an approach for the CoVR task. \\
\textbf{Query-specific Contextual Information Preservation}: As discussed earlier, compositional video retrieval (CoVR) relies on reducing the domain gap between the visual input and the change text. This is typically achieved by leveraging an image-grounded text encoder, where the cross-attention layers are trained using the changing text and embedding from a frozen visual encoder \cite{ventura2023covr}. As the training occurs within the image-grounded text encoder only, the multi-modal representation is likely to get predominately biased towards the change text. As a result, it looses the context of the query sample which is essential for the task.  

In this work, we argue that such a query-specific contextual information can be incorporated in the image-grounded text encoder through detailed language descriptions of these query videos; the red liquid is for printing machine not immediately visible in the input, the white tulip stands out from the crowd of colorful red and yellow flowers, and the sun is setting over the ocean and behind the clouds (see \cref{fig:language descriptions demo}). 
 
Therefore, the correspondence between embedding of the visual input ($q$) and its corresponding detailed description ($d$) results in an enhanced vision-text representation $f(q,d)$ of $q$, thereby ensuring a contextualized understanding of the query video. 
Further, the complementary nature of detailed descriptions aids in reducing the domain gap between the input query and the modification text by establishing correspondence between the detailed description of the input query and the modification text, as $f(d, t)$. Thus, we seek to improve CoVR by minimizing the following objective:
\begin{equation}
\label{eq:main_eq}
   v^* = \underset{v\in V}{\arg\max} \;\;\; \mathcal{L}\left( \Tilde{f}(q,d,t), \;\; g(v)\right),
\end{equation}
\begin{equation}
\Tilde f(q,d,t) = f(q,t) + f(q,d) + f(e(d),t).
\label{eq:added_emb}
\end{equation}
where, $q$ and $d$ represent input query (image/video) and its corresponding language description, $t$ is the desired modification text, $\Tilde{f}(q,d,t)$ is the pairwise summation of individual correspondence embeddings and $\mathcal{L}$ is a similarity-based loss.\\
\noindent\textbf{Learning Discriminative Embeddings for Alignment: } In the CoVR task, the model is desired to learn to align its output with the target video after mixing the change text with the query video. Instead of only representing the target video in the latent space through a visual embedding \cite{ventura2023covr}, a multiple discriminative embedding of vision, text, and vision-text is expected to provide better alignment due to complementary target video representation. 

\noindent\textbf{Overall Architecture: } Figure~\ref{fig:architecture} presents our proposed architecture comprising three inputs: the reference video, the text corresponding to the change, and the detailed video description. Compared to the baseline framework, the focus of our design is to effectively align the joint multi-modal embedding, comprised of the three inputs ($\Tilde{f}(q,d,t)$), with the target video database to achieve enhanced contextual understanding during training for composed video retrieval. Within our proposed framework, we first process the reference video and its description using pre-trained \cite{li2023blip} image encoder $g$ and text encoder $e$ to produce their latent embedding of the same dimension as, $q \in \mathbb{R}^{m}$ and $d \in \mathbb{R}^{m}$. We use the same multi-modal encoder $f$, as in the baseline \cite{ventura2023covr}. This multi-model encoder takes  the  visual embeddings from a pre-trained visual encoder $g$ along with tokenized textual inputs and produces a multi-modal embedding. Given the tokenized change text $t$, and embeddings of the reference video and its descriptions, $q$ and $d$, we fuse any two inputs at a time using the multi-modal encoder $f$ comprising of cross-attention layers, to produce joint multi-modal embeddings ($\Tilde{f}(q,d,t)$), as shown in \cref{fig:architecture}. The input query video and its corresponding description are processed by visual encoder $g$ and text encoder, respectively. It is worth mentioning that the only difference between the text encoder $e$ and the multi-modal encoder $f$ are cross attention layers. In other words, if we remove the cross attention layers from multi-modal encoder $f$, it converts to text-only encoder $e$. We use the text encoder $e$ to process the language descriptions of an input video.

Within the proposed framework, we only train the multi-modal encoder $f$ whereas the image and text encoders remain frozen.  During training, we provide the change text $t$ and the visual query embeddings $q$ to the encoder $f$ for obtaining the multi-model embeddings $f(q,t)$ corresponding to the change text $t$. As shown in \cref{fig:architecture}, here grounding occurs via cross-attention between $q$ and $t$. In a similar manner, we obtain an enhanced contextualized multi-model representation of embeddings of input video $q$ and its tokenized description  $d$ from  $f$ as $f(q,d)$. As a final step, we provide the change text $t$ to the text encoder $f$, and ground it with the embedding of description $e(d)$ to obtain $f(e(d),t)$. Consequently, we combine these grounded embeddings in a pairwise summation manner as shown in~\cref{eq:added_emb} to obtain the joint multi-model embeddings $\Tilde{f}(q,d,t)$. These joint multi-model embeddings are then utilized to retrieve the target video from the database. In order to train the multi-modal encoder $f$, we employ hard-negative contrastive loss \cite{ventura2023covr,hn_nce} between $\Tilde{f}(q,d,t)$ and the target database, as shown in~\cref{fig:architecture}. The loss is as follows, 
\begin{align}
\label{eq:hn_nce}
\mathcal{L}= -\sum_{i \in \mathcal{B}}\text{log}\left(\frac{e^{S_{i,i}/\tau}}{\alpha \cdot {e^{S_{i,i}/\tau}} + \sum_{j \neq i}{e^{S_{i,j}/\tau}w_{i,j}} }\right) \nonumber \\ - \sum_{i \in \mathcal{B}} \text{log}\left(\frac{e^{S_{i,i}/\tau}}{\alpha \cdot {e^{S_{i,i}/\tau}} + \sum_{j \neq i}{e^{S_{j,i}/\tau}w_{j,i}} }\right)
\end{align}
where $\alpha$ is set to 1 and temperature $\tau$ is set to $0.07$ as in~\cite{hn_nce}, $S_{i,j}$ is the cosine similarity between the joint multi-modal embedding $\Tilde{f}(q_i,d_i,t_i)$ and the corresponding target video $g(v_i)$, $w_{i,j}$ is set as in \cite{hn_nce} with $\beta=0.5$, and $\mathcal{B}$ is the batch size. Next, we describe how to effectively utilize the recent multi-modal conversation models \cite{zhu2023minigpt} to obtain query-specific detailed language descriptions for composed video retrieval. 
%

\subsection{Query-specific Language Descriptions} 
\begin{figure}[!t]
  \centering
   \includegraphics[width=\linewidth]{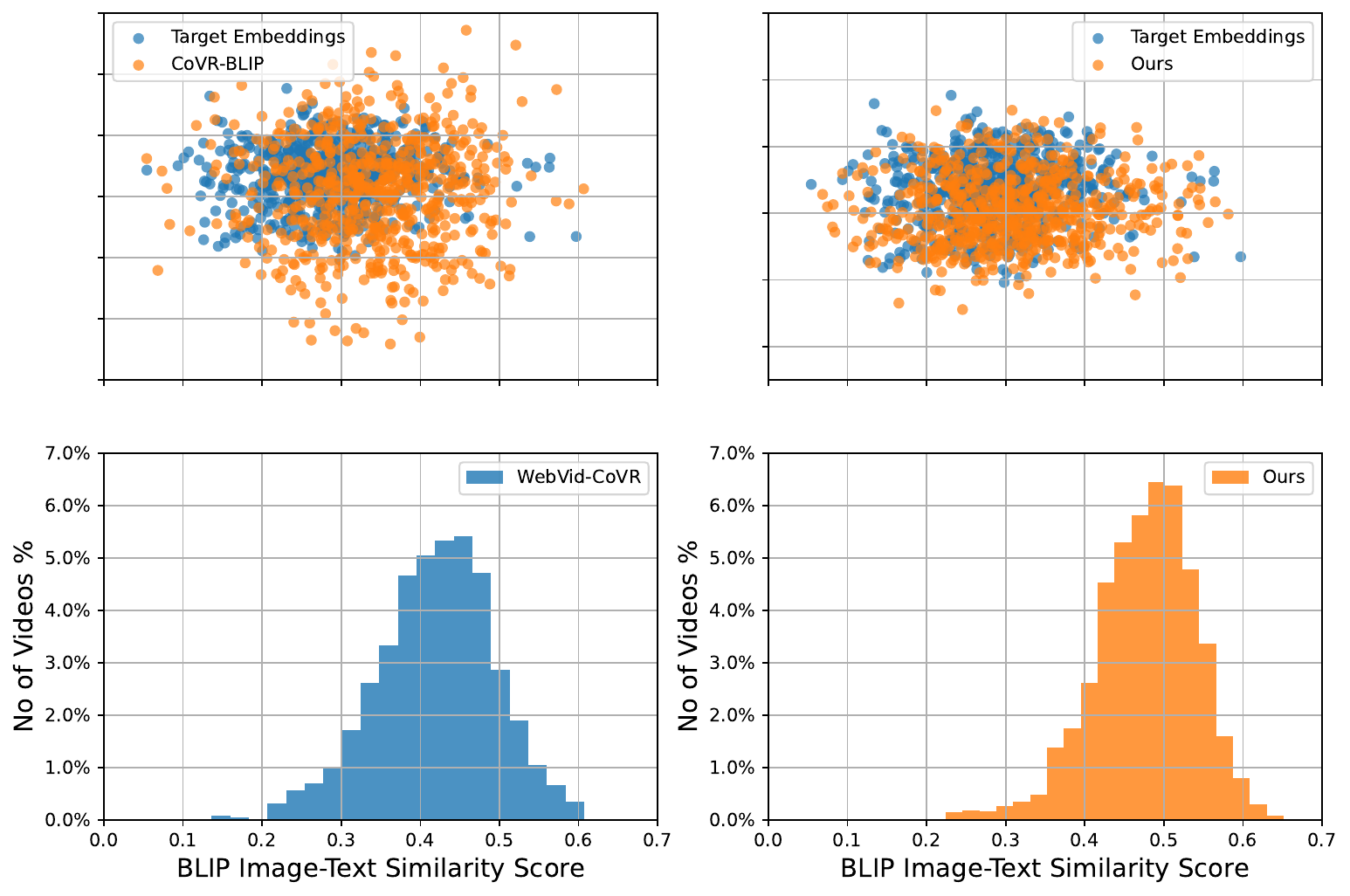}
   \caption{\textbf{First row:} Comparison between the baseline and our approach in terms of proximity of the  output embedding with the target videos on WebVid dataset. Here, each data sample represents the projection of the embedding from $\mathbb{R}^m$ to $\mathbb{R}^2$. Our joint multi-modal embeddings leveraging the language information are closer to the target embeddings, compared to the baseline embedding utilizing only the visual input. \textbf{Second row:} the cosine similarity between video embeddings and the WebVid dataset captions (on the left), compared to the similarity between video embeddings and our generated textual descriptions (on the right). Here, the Y-axis corresponds to the number of videos whereas the X-axis denotes the cosine similarity. Our approach utilizing the generated descriptions achieves better alignment with the video embeddings. 
   }
   \label{fig:embeds}
\end{figure}

\noindent In order to obtain the video descriptions, we employ a recent open-source multi-modal conversation model~\cite{zhu2023minigpt}. Generally, multi-modal conversation models learn alignment between a pretrained large language model such as, Vicunna~\cite{vicuna2023} and a pretrained vision encoder of vision language model such as, CLIP \cite{radford2021learning} or BLIP \cite{li2022blip}. This alignment enables these multi-modal conversation models to reason and contextualize a given visual input. Since these are image models and for our case of video inputs, we sample the middle frame of the video and generate its detailed description using a multi-modal conversation model by prompting the model with "Describe the input image in detail". We further remove the noise within these descriptions by removing the manually curated unnecessary symbols, tokens, or special characters. Further, these models can hallucinate about a given visual sample. To identify hallucinated descriptions, we first measure the lower bound of cosine similarity between default WebVid captions \cite{bain2021frozen} and visual inputs within BLIP latent space to identify a hallucination threshold. We then discard those descriptions, where the cosine similarity between our generated description and the visual input is lower than the hallucination threshold. 
Consequently, the resulting enriched descriptions are better aligned with the videos (\cref{fig:embeds}). As discussed earlier, the base framework \cite{ventura2023covr} only aligns the input video with the target video database. To further enhance the alignment of our joint multi-modal $\Tilde{f}(q,d,t)$, we introduce multiple target datasets as explained next.

\begin{table*}[!t]
\setlength\tabcolsep{9pt}
  \centering
    \scalebox{0.8}[0.8]{
    \begin{NiceTabular}{clclccc|cccc}
    \toprule
     & & Training  &  &  &  &   &  \multicolumn{4}{c}{Recall@K}  \\
     & Model &  WebVid-CoVR & Input Modalities & Fusion & Backbone &  Frames & R@1 & R@5 & R@10 & R@50 \\
    \midrule
    1 & Random & & - & - & - & - & 0.08 & 0.23 & 0.35 & 1.76 \\
    2 & CoVR-BLIP \cite{ventura2023covr} & \ding{56} & Text & - & BLIP & - & 19.68 & 37.09 & 45.85 & 65.14 \\
    3 & CoVR-BLIP \cite{ventura2023covr} & \ding{56} & Visual & - & BLIP & 15 & 34.90 & 59.23 & 68.04 & 85.95 \\
    4 & CoVR-BLIP \cite{ventura2023covr} & \ding{56} & Visual + Text & Avg & CLIP & 15 & 44.37 & 69.13 & 77.62 & 93.00 \\
    5 & CoVR-BLIP \cite{ventura2023covr} & \ding{56} & Visual + Text & Avg & BLIP & 15 & 45.46 & 70.46 & 79.54 & 93.27 \\
    \rowcolor{orange!15}
    6 & \textbf{Our Approach} & \ding{56}  & Visual + Text & Avg & BLIP & 15 & \textbf{47.52} & \textbf{72.18} & \textbf{82.37} & \textbf{95.06} \\
    \midrule
    7 & CoVR-BLIP \cite{ventura2023covr} & \ding{56} & Visual + Text & CA & BLIP & 15 & 15.85 & 32.79 & 40.3 & 58.33 \\ 
    \rowcolor{orange!15}
    8 & \textbf{Our Approach} & \ding{56}  & Visual + Text & CA & BLIP & 15 & \textbf{20.85} & \textbf{41.2} & \textbf{50.2} & \textbf{72.1} \\ 
    \midrule
    9 & CoVR-BLIP \cite{ventura2023covr} & \ding{52}  & Text & - & BLIP & - & 23.67 & 45.89 & 55.13 & 77.03 \\
    10 & CoVR-BLIP \cite{ventura2023covr}& \ding{52}  & Visual & - & BLIP & 15 & 38.89 & 64.98 & 74.02 & 92.06 \\
    11 & CoVR-BLIP \cite{ventura2023covr}& \ding{52}  & Visual + Text & MLP & CLIP & 1 & 50.55 & 77.11 & 85.05 & 96.06 \\
    12 & CoVR-BLIP \cite{ventura2023covr}& \ding{52}  & Visual + Text & MLP & BLIP & 1 & 50.63 & 74.8 & 83.37 & 95.54 \\
    13 & CoVR-BLIP \cite{ventura2023covr}& \ding{52}  & Visual + Text & CA & BLIP & 1 & 51.80 & 78.29 & 85.84 & 97.07 \\
    14 & CoVR-BLIP \cite{ventura2023covr}& \ding{52}  & Visual + Text & CA & BLIP & 15 & 53.13 & 79.93 & 86.85 & 97.69 \\
    \rowcolor{orange!15}
    15 & \textbf{Our Approach} & \ding{52} & Visual + Text & CA & BLIP & 15 & \textbf{60.12} & \textbf{84.32} & \textbf{91.27} & \textbf{98.72} \\
    \bottomrule
    \end{NiceTabular}}
    \vspace{-0.5em}
    \caption{\textbf{Baseline comparison on the WebVid-CoVR test set}. Without training on the WebVid-CoVR and using averaging as fusion, our approach (row 6) achieves a gain of xx over the baseline (row 5). A consistent improvement in performance is also obtained over the baseline (row 7 vs. row 8) when using cross-attention (CA) as a fusion scheme. The performance is improved when performing training on the WebVid-CoVR training set. Using the same input modalities, fusion scheme, backbone, and frames, our approach (row 15) achieves a significant gain of 6.9\% in terms of Recall@K=1 over the baseline \cite{ventura2023covr} (row 14). Best results are in bold.
    }

  \label{tab:covr_results}
  \vspace{-1em}
\end{table*}

\subsection{Enhancing Diversity in Target Database}
\label{sec:target_datasets}
The proposed method takes three inputs (video, modification text, and video description) and three target databases to train the model. The first target database is based on the visual embedding of input videos generated by a pretrained vision encoder of BLIP-2~\cite{li2023blip}. Our second target database is based on multi-model embeddings derived from the pretrained multi-modal encoder of BLIP-2~\cite{li2023blip}. The final target database is based on a text-only embedding of the video description generated by the pretrained BLIP-2~\cite{li2023blip} text encoder. We use these additional databases only during training time to compute the hard negative contrastive loss between our joint multi-model embeddings and target datasets. 

\noindent\textbf{Overall Loss Formulation: }
For a given batch $\mathcal{B}$, we formulate hard negative contrastive loss for each of our three target databases as follows,

\begin{equation}
\label{eq:adding_loss}
   \mathcal{L}_{contr}  = \lambda * \mathcal{L}_{ve} + \mu * \mathcal{L}_{mme} + \delta * \mathcal{L}_{te}, 
\end{equation}
where, $\mathcal{L}_{ve}$, $ \mathcal{L}_{mme}$, and  $ \mathcal{L}_{te}$ are the contrastive loss represented by \cref{eq:hn_nce}. We compute the similarity of $\Tilde{f}(q_i,d_i,t_i)$ with the corresponding target video embedding $g(v_i)$, target multi-modal embedding $f(g(v_i), d_{v_i})$, and the target text embedding $d_{v_i}$ for  $\mathcal{L}_{ve}$, $ \mathcal{L}_{mme}$, and  $ \mathcal{L}_{te}$, respectively. $\lambda$, $\mu$, and $\delta$ are learnable parameters that scale the weightage of each loss during training.

\noindent \textbf{Inference: } During inference, for the 3 given inputs: reference video, description and change text,  we first process the reference video and its description using pre-trained frozen image encoder $g$ and text encoder $e$ to produce their latent embedding. The change text is simply tokenized as shown in Figure~\ref{fig:architecture}. We use the multi-modal encoder $f$ and gather the multi-model embeddings from 2 inputs at a time such as $f(q,t)$, $f(q,d)$ and $f(e(d),t)$. Consequently, we simply do the pairwise addition of three (3) multi-model embeddings to produce joint multi-modal embeddings $\Tilde{f}(q,d,t)$ for target video retrieval. Note that, this pairwise addition allows us to use any combination of inputs as illustrated in ablative analysis (refer \cref{tab:Effect of Input Combinations}). Finally, similar to CoVR~\cite{ventura2023covr} the target videos are retrieved by mapping the similarity between the joint multi-model embeddings $\Tilde{f}(q,d,t)$ and the visual embedding database {${g(V)}$}.

\begin{figure*}[!t]
  \centering
   \includegraphics[width=\textwidth]{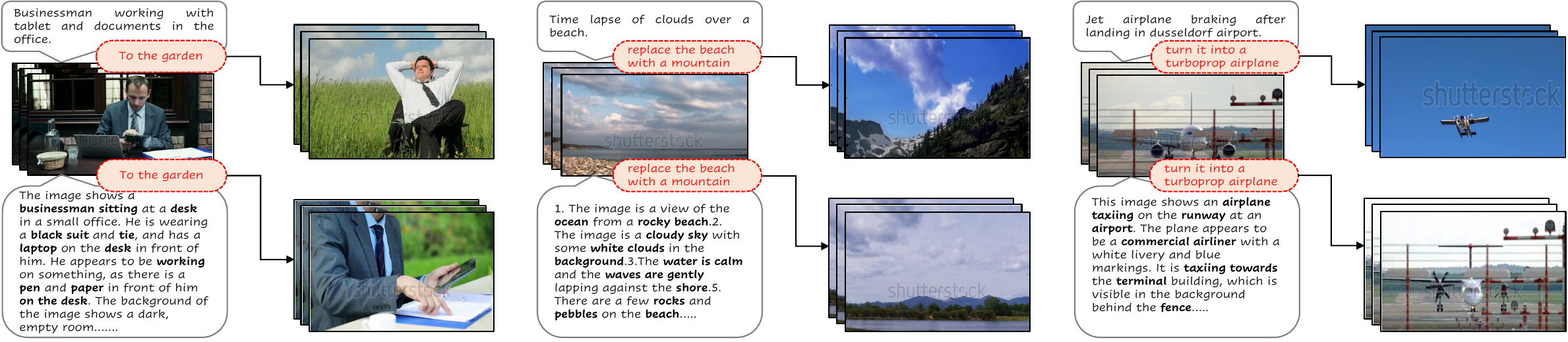}
   \caption{
   Qualitative Comparison between default WebVid-CoVR short captions (top row) with our generated detailed descriptions (bottom row) within our framework. The change text is highlighted in red and the text (default short captions in top row, detailed description in bottom row) are highlighted in black. Here in all three examples from CovR-Vid testset, we observe that the default WebVid-CoVR short captions struggle to fully preserve the contextual information in the retrieved target video (top row). In comparison, our approach leveraging detailed descriptions is able to correctly retrieve the target video with most relevant contextual match with reference video (bottom row). For instance, keeping person working while putting him beside garden in \textit{video-1}, keeping the sea while replacing the beach with mountains in\textit{ video-2} and keeping the turboprop airplane on airport behind fence in \textit{video-3}. Best viewed zoomed in. Additional examples are in the suppl.   
   }
   \label{fig:qualitative_figure}
\end{figure*}

\section{Experiments}

\subsection{Experimental Setup and Protocols}
\textbf{Dataset for Composed Video Retrieval (CoVR):} We evaluate our approach on the recently introduced WebVid-CoVR dataset~\cite{ventura2023covr}.  The training set of WebVid-CoVR consists of triplets (input video, change text, and target video) and is generated synthetically, whereas the test set is manually curated using the model in the loop. The change text within a triplet is generated by comparing captions of the input and target videos using an LLM. It represents the differences between the input and target videos. The WebVid-CoVR training set consists of 131K distinct videos and 467K distinct change texts. One video is associated with each of the 12.7 triplets and the average change text length is 4.8 words. WebVid-CoVR also includes validation and test sets gathered from the WebVid10M corpus. In the validation set there are 7K triplets, whereas in the test set there are 3.2K triplets that have been manually curated to ensure high quality.

\noindent\textbf{Datasets for Composed Image Retrieval (CoIR):} We use CIRR~\cite{cirr} and FashionIQ~\cite{fashioniq} benchmarks for composed image retrieval. CIRR~\cite{cirr} consists of manually annotated open-domain natural image and change text pairs with (36.5K, 19K) distinct pairs. The data distribution of this image and change text pairs is around (28.2K,16.7K), (41.8K, 22.6K), (41.5K, 21.8K) for training, testing, and validation set, respectively. The FashionIQ~\cite{fashioniq} dataset consists of images of fashion products in three categories: Shirts, Dresses, and Tops/Tees. The reference query and target image are paired based on their category. The corresponding change text is manually annotated. This dataset consists of (30K, 40.5K) images and change text pairs queries annotated on 40.5K distinct images. The data distribution of this image and change text pairs is around (18K, 45.5K), (60.2K, 15.4K) for training, testing, and validation, respectively. \\
\noindent \textbf{Evaluation Metrics: } We follow standard evaluation protocol for the composed image as well video retrieval from ~\cite{cirr, ventura2023covr}. We report the retrieval results using recall values at rank 1, 5, 10, 50. Recall at rank k (R@k) denotes the number of times the correct retrieval occurred among the top-k results. \\ 
\noindent \textbf{Implementation Details: } We use a multi-modal conversational model \cite{zhu2023minigpt} to generate the visual descriptions. As discussed earlier, we built our approach on the recent CoVR-BLIP~\cite{ventura2023covr} and use the same components without adding any \emph{additional parameters}. We use ViT-L~\cite{dosovitskiy2020transformers} as the frozen vision encoder $g$, which is pretrained for text-image retrieval on COCO~\cite{coco}. The frozen text encoder $e$ is from BLIP-2~\cite{li2023blip} without cross-attention with pretrained weights of BERT$_{base}$~\cite{devlin2018bert}. We train our model for 20 epochs with a batch size of 1024 (256 batch size per device) with an initial learning rate of $1e-5$. For a fair comparison, we report the results of our baseline CoVR-BLIP~\cite{ventura2023covr} in the same settings. For transfer learning on CoIR, we fine-tuned the model on the FashionIQ dataset for 6 epochs. We use a batch size of 2048/1024 and an initial learning rate of $1e-4$. After training, our learnable parameters $\lambda$, $\mu$ and $\delta$ for scaling the weightage of each loss was optimized based on validation set with values $0.83$, $0.08$ and $0.07$.  We use  four NVIDIA A100 GPUS for all the experiments.

\subsection{Composed Video Retrieval}
\noindent\textbf{Baseline Comparison: }
We present a baseline comparison in \cref{tab:covr_results}. Compared to the baseline CoVR-BLIP~\cite{ventura2023covr}, our approach achieves consistent improvement in performance across different recall rates. Without training on WebVid-CoVR benchmark, our approach achieves a significant gain of 5.0\% in terms of Recall@K=1 and 13.8\% in terms of Recall@K=50. When conducting training on WebVid-CoVR dataset, our approach achieves a significant improvement of ~7\% in terms of Recall@K=1. \\
\noindent\textbf{Ablation Study: } We first analyze the \textit{impact of inputs} on CoVR performance. Here, we train our model as described in \cref{sec:Method}. We freeze our model and study the effect of inputs: reference video, descriptions, and change text with their different combinations for CoVR during inference (\cref{tab:Effect of Input Combinations}) on WebVid-CoVR test set. The performance increases as we replace the input from change text to descriptions to the reference video. As soon as the video and descriptions are provided, the performance improves. This shows that the detailed descriptions provide additional information that complements the input video. Since modification text is not part of the input, the model did not have instructions regarding how to change the composition of the input video and behaves as a plain retrieval task. Further, we obtain superior results after providing modification text along with the detailed descriptions and input video.

\begin{table}[!t] %
    \caption{\textbf{The impact of Inputs on our model performance on WebVid-CoVR testset.} 
     The best performance is obtained when using all inputs (video, detailed description and change text), indicating the complementary nature of videos and their detailed language descriptions. Best results in bold.
    }
    \label{tab:Effect of Input Combinations}
    \centering
    \resizebox{1\linewidth}{!}{
    \begin{tabular}{ccc|cccc}
    \toprule
    \multicolumn{3}{c|}{Input} & \multicolumn{4}{c}{Recall@K}\\
        Video ($q$) & description ($d$) & change text ($t$) & R@1 & R@5 & R@10 & R@50 \\
        \toprule
         \ding{56} & \ding{56} & \ding{51} & 26.95 & 51.25 & 62.19 & 83.24 \\
         \ding{56} & \ding{51} & \ding{56}  & 39.92 & 65.62 & 76.21 & 92.23 \\
         \ding{51} & \ding{56} & \ding{56} & 40.51 & 66.56 & 76.95 & 92.66 \\
         \ding{51} & \ding{51} & \ding{56} & 42.19 & 69.06 & 78.95 & 95.0 \\
         \ding{56} & \ding{51} & \ding{51} & 45.51 & 73.12 & 82.7 & 95.78 \\
         \ding{51} & \ding{56} & \ding{51}  & 56.26 & 81.46 & 88.97 & 98.0  \\
        \midrule
        \rowcolor{orange!15}
         \ding{51} & \ding{51} & \ding{51} & \textbf{60.12} & \textbf{84.32} & \textbf{91.27} & \textbf{98.72} \\
        \bottomrule
    \end{tabular}
    }
\end{table}

\begin{table}[!t] %
    \caption{\textbf{ The impact of target datasets on our model performance on WebVid-CoVR test set.} We study the impact of different configurations of losses ($\mathcal{L}_{ve}$, $\mathcal{L}_{mme}$, $\mathcal{L}_{te}$) during training. The best results are obtained with all loss terms, indicating the importance of diversity in target datasets. Best results are in bold.    
    }
    \label{tab:loss_change}
    \centering
    \setlength{\tabcolsep}{12pt}
    \resizebox{1\linewidth}{!}{
    \begin{tabular}{ccc|cccc}
    \toprule
    \multicolumn{3}{c|}{Training Loss} & \multicolumn{4}{c}{Recall@K} \\
        $\mathcal{L}_{ve}$ & $\mathcal{L}_{mme}$ & $\mathcal{L}_{te}$ & R@1 & R@5 & R@10 & R@50 \\
        \toprule
        \ding{56} & \ding{56} & \ding{51} & 33.79 & 65.47 & 77.93 & 94.65 \\
        \ding{56} & \ding{51} & \ding{56} & 42.97 & 69.49 & 78.71 & 93.2 \\
        \ding{51} & \ding{56} & \ding{56} & 58.37 & 83.72 & 89.79 & 98.16 \\
        \ding{56} & \ding{51} & \ding{51} & 58.94 & 83.86 & 89.76 & 98.64 \\
        \ding{51} & \ding{51} & \ding{56} & 59.12 & 84.18 & 90.36 & 98.54 \\  
        \midrule
        \rowcolor{orange!15}
        \ding{51} & \ding{51} & \ding{51} & \textbf{60.12} & \textbf{84.32} & \textbf{91.27} & \textbf{98.72} \\
        \bottomrule
    \end{tabular}
    }
\end{table}

Next, we study in Tab.~\ref{tab:loss_change} the \textit{effect of different target datasets}: visual embeddings, multi-modal embeddings, and text-only embedding as explained in \cref{sec:target_datasets}. This implies that we use any or different combination of the three contrastive losses introduced in \cref{eq:adding_loss} ($\mathcal{L}_{ve}$, $\mathcal{L}_{mme}$, $\mathcal{L}_{te}$). We use all three inputs at inference for CoVR. As a result of training only with $\mathcal{L}_{te}$, we observe an improvement indicating that  $\mathcal{L}_{te}$ plays a role in refining our joint multi-modal embedding. Introducing $\mathcal{L}_{mme}$  as a training loss further enhances recall rates, compared to $\mathcal{L}_{te}$, emphasizing its positive impact on the model's ability to retrieve relevant instances. When $\mathcal{L}_{ve}$ is employed as a training loss, the performance improves across all metrics. Similarly, the combination of these losses gives further improvement in performance, indicating their complementing nature.

\begin{table}[!t] %
\setlength\tabcolsep{9pt}
    \caption{\textbf{State-of-the-art comparison on CIRR test set}. Our approach achieves consistent improvement in performance on both transfer learning (row 1-12) and zero-shot settings (row 13-20). On the challenging zero-shot setting and using the same pretraining WebVid-CoVR dataset, our approach achieves an absolute gain of 1.6\% in terms of Recall@K=1 over ~\cite{ventura2023covr}. Best results are in bold. 
    }
    \resizebox{1\linewidth}{!}{
    \begin{tabular}{clc|ccc|cc}
        \toprule
         & & Pretrain  & \multicolumn{3}{c|}{Recall@K} & \multicolumn{2}{c}{$\text{R}_{\text{subset}}$@K} \\
         & Method & Data & K=1 & K=10 & K=50 & K=1  & K=3 \\ 
        \midrule
        \multirow{12.2}{*}{
        \begin{tabular}[c]{@{}c@{}}
        \rotatebox{90}{\small{Train CIRR}}\end{tabular}}
         & TIRG \cite{tirg} & - & 14.61  & 64.08 & 90.03 & 22.67  & 65.14 \\
         & TIRG+LastConv \cite{tirg}$\dagger$ & - & 11.04  & 51.27 & 83.29 & 23.82  & 64.55 \\
         & MAAF \cite{MAAF} & - & 10.31  & 48.30 & 80.06 & 21.05  & 61.60 \\
         & MAAF-BERT \cite{MAAF}$\dagger$ & - & 10.12  & 48.01 & 80.57 & 22.04  & 62.14 \\
         & MAAF-IT \cite{MAAF} & - & \textcolor{white}{0}9.90  & 48.83 & 80.27 & 21.17  & 60.91 \\
         & MAAF-RP \cite{MAAF} & - & 10.22  & 48.68 & 81.84 & 21.41  & 61.60 \\
         & ARTEMIS \cite{ARTEMIS} & - & 16.96  & 61.31 & 87.73 & 39.99  & 75.67 \\
         & CIRPLANT \cite{cirr} & - & 19.55  & 68.39 & 92.38 & 39.20  & 79.49 \\
         & LF-BLIP \cite{cclip,levy2023case} & - & 20.89  & 61.16 & 83.71 & 50.22  & 86.82 \\
         & CompoDiff~\cite{gu2023compodiff} & \ding{51} & 22.35  & 73.41 & 91.77 & 35.84  & 76.60 \\ %
         & Combiner \cite{cclip} & - & 33.59  & 77.35 & 95.21 & 62.39  & 92.02 \\
         & CASE~\cite{levy2023case} & - & 48.00  & 87.25 & \textbf{97.57} & 75.88 & \textbf{96.00} \\
         & CASE~\cite{levy2023case} & \ding{51} & {49.35}  & 88.75 & 97.47 & 76.48  & 95.71 \\ 
         & \ourM ~\cite{ventura2023covr}  & - & 48.84  & 86.10 & 94.19 & 75.78 & 92.80 \\
         & Ours  & -  & 49.18 & 87.06 & 94.72 & 75.66  & 93.16 \\ 
         & \ourM ~\cite{ventura2023covr}  & \ding{51} & 49.69  & 86.77 & 94.31 & 75.01 & 93.16 \\
        \rowcolor{orange!15} \cellcolor{white}
         & Ours & \ding{51} & \textbf{51.03} & \textbf{88.93} & \textbf{97.53} & \textbf{76.51} & \textbf{95.76} \\ 
        \midrule
        \multirow{9.2}{*}{\begin{tabular}[c]{@{}c@{}} 
        \rotatebox{90}{\small{Zero Shot}}\end{tabular}} 
         & Random$\dagger$  & - & \textcolor{white}{0}0.04 & \textcolor{white}{0}0.44 & \textcolor{white}{0}2.18 & 16.67 & 50.00 \\
         & CompoDiff \cite{gu2023compodiff} & \ding{51} & 19.37 & 72.02 & 90.85 & 28.96 & 67.03 \\ %
         & Pic2Word~\cite{saito2023pic2word} & \ding{51} & 23.90  & 65.30 & 87.80 & - & - \\
         & CASE~\cite{levy2023case} & \ding{51} & 30.89  & 73.88 & 92.84 & 60.17  & 90.41 \\
         & CASE~\cite{levy2023case} & \ding{51} & 35.40  & 78.53 & 94.63 & 64.29  & 91.61 \\ 
         & \ourM ~\cite{ventura2023covr} & - & 19.76  & 50.89 & 71.64 & 63.04  & 89.37 \\
         & Ours  & - & 21.34  & 52.37 & 74.92 & 64.66  & 90.87 \\ 
         & \ourM ~\cite{ventura2023covr} & \ding{51}  & 38.48 & 77.25 & 91.47 & 69.28  & 91.11 \\
        \rowcolor{orange!15}\cellcolor{white}
         & Ours & \ding{51} & \textbf{40.12} & \textbf{78.86}  & \textbf{94.69} & \textbf{70.47}  & \textbf{92.12} \\ 
    \bottomrule
\end{tabular}
}
\label{tab:cirr_results}
\end{table}

\begin{table}[!t]
    \caption{\textbf{State-of-the-art comparison on FashionIQ val. set}. Our method obtains favorable results on both transfer learning (row 1-19) and zero-shot (20-25) settings. On the challenging zero-shot setting and using same pretraining, our method obtains an absolute gain of 2.6\% (average over three classes: \textit{Dress}, \textit{Shirt} and \textit{Toptee}) in terms of Recall@R=10 over ~\cite{ventura2023covr}. Best results are in bold.
    }
    \resizebox{1\linewidth}{!}{
    \begin{tabular}{clc|cc|cc|cc}
        \toprule
        & & Pretrain & \multicolumn{2}{c}{Dress} & \multicolumn{2}{c}{Shirt} & \multicolumn{2}{c}{Toptee} \\
        & Method & Data & R@10 & R@50 & R@10 & R@50 & R@10 & R@50  \\
        \midrule
        \multirow{18.2}{*}{
        \begin{tabular}[c]{@{}c@{}}
        \rotatebox{90}{\small{Train FashionIQ}} \end{tabular}}
        & JVSM \cite{JVSM} & - & 10.70 & 25.90 & 12.00 & 27.10 & 13.00 & 26.90  \\
        & CIRPLANT \cite{cirr} & - & 17.45 & 40.41 & 17.53 & 38.81 & 61.64 & 45.38  \\
        & TRACE \cite{TRACE} & - & 22.70 & 44.91 & 20.80 & 40.80 & 24.22 & 49.80  \\
        & VAL w/GloVe \cite{VAL_IR} & - & 22.53 & 44.00 & 22.38 & 44.15 & 27.53 & 51.68  \\
        & MAAF \cite{MAAF} & - & 23.80 & 48.60 & 21.30 & 44.20 & 27.90 & 53.60  \\
        & CurlingNet \cite{CurlingNet} & - & 26.15 & 53.24 & 21.45 & 44.56 & 30.12 & 55.23  \\
        & RTIC-GCN \cite{RTIC} & - & 29.15 & 54.04 & 23.79 & 47.25 & 31.61 & 57.98  \\
        & CoSMo\cite{Lee_2021_CVPR_cosmo} & - & 25.64 & 50.30 & 24.90 & 49.18 & 29.21 & 57.46  \\
        & ARTEMIS\cite{ARTEMIS} & - & 27.16 & 52.40 & 21.78 & 43.64 & 29.20 & 53.83  \\
        & DCNet\cite{Kim_Yu_Kim_Kim_2021_dcnet} & - & 28.95 & 56.07 & 23.95 & 47.30 & 30.44 & 58.29 \\
        & SAC \cite{SAC} & - & 26.52 & 51.01 & 28.02 & 51.86 & 32.70 & 61.23  \\
        & FashionVLP\cite{FashionVLP} & - & 32.42 & 60.29 & 31.89 & 58.44 & 38.51 & 68.79  \\
        & LF-CLIP \cite{cclip} & - & 31.63 &  56.67 & 36.36 &  58.00 &  38.19 & 62.42  \\
        & LF-BLIP~\cite{cclip,levy2023case} & - & 25.31 & 44.05  &  25.39 & 43.57  &  26.54 & 44.48   \\
        & CASE~\cite{levy2023case} & \ding{51} & \textbf{47.44} & 69.36 & 48.48 & \textbf{70.23} & {50.18} & {72.24}  \\ 
        & \ourM ~\cite{ventura2023covr} & -   & 43.51 & 67.94 & 48.28 & 66.68 & 51.53 & 73.60  \\
        & Ours & - & 44.39 & 68.86 & 49.17 & 67.54 & 52.47 & 74.28  \\
        & \ourM ~\cite{ventura2023covr} & \ding{51} & 44.55 & 69.03 & 48.43 & 67.42 & 52.60 & 74.31  \\
        \rowcolor{orange!15} \cellcolor{white}
        & Ours & \ding{51} & 46.12 & \textbf{69.52} & \textbf{49.61} & 68.88 & \textbf{53.79} & \textbf{74.74}  \\
        \midrule
        \multirow{5.2}{*}{\begin{tabular}[c]{@{}c@{}}  
        \rotatebox{90}{\small{Zero Shot}}\end{tabular}} 
        & Random & - & \textcolor{white}{0}0.26 & \textcolor{white}{0}1.31 & \textcolor{white}{0}0.16 & \textcolor{white}{0}0.79 & \textcolor{white}{0}0.19 & \textcolor{white}{0}0.95  \\
        & Pic2Word~\cite{saito2023pic2word} & \ding{51} & 20.00 & 40.20 & 26.20 & 43.60 & 27.90 & 47.40 \\
        & \ourM ~\cite{ventura2023covr} & - & 13.48 & 31.96 & 16.68 & 30.67 & 17.84 & 35.68  \\
        & Ours & - & 15.24 & 34.12 & 18.36 & 32.54 & 19.56 & 37.54 \\
        & \ourM ~\cite{ventura2023covr} & \ding{51} &  21.95 & 39.05 & 30.37 & 46.12 & 30.78 & 48.73   \\
        \rowcolor{orange!15} \cellcolor{white}
        & Ours & \ding{51} &  \textbf{24.57} & \textbf{40.93} & \textbf{33.12} & \textbf{48.42} & \textbf{33.16} & \textbf{50.24}  \\
    \bottomrule
\end{tabular}
}
\label{tab:fashioniq_results}
\end{table}

Lastly, we analyze the impact of \textit{detailed language descriptions} on CoVR performance. To this end, we train our model as described in \cref{sec:Method} and freeze our model to study the impact of video description quality for CoVR. We compare our detailed description with short WebVid captions (Tab.~\ref{tab:webvid_captions}) and observe that the performance is inferior when using the default WebVid captions, possibly because they do not provide richer context compared to detailed descriptions. \cref{fig:qualitative_figure} further shows this comparison on example video samples from WebVid-CoVR test set.

\begin{table}[!t] %
    \caption{\textbf{The impact of detailed descriptions on our model performance on WebVid-CoVR test set.} Our approach leveraging detailed descriptions achieves consistent improvement in performance, compared to the default captions. Best results are in bold.
    }
    \label{tab:webvid_captions}
    \centering
    \setlength{\tabcolsep}{10pt}
    \resizebox{1\linewidth}{!}{
    \begin{tabular}{l|cccc}
    \toprule
        Our Model & R@1 & R@5 & R@10 & R@50 \\
        \toprule
         using webvid captions & 58.23 & 83.31 & 90.08& 98.05 \\ 
        \rowcolor{orange!15}
         using our detailed descriptions & \textbf{60.12} & \textbf{84.32} & \textbf{91.27} & \textbf{98.72} \\
        \bottomrule
    \end{tabular}
    }
\end{table}

\begin{figure}[!t]
  \centering
   \includegraphics[width=\linewidth]{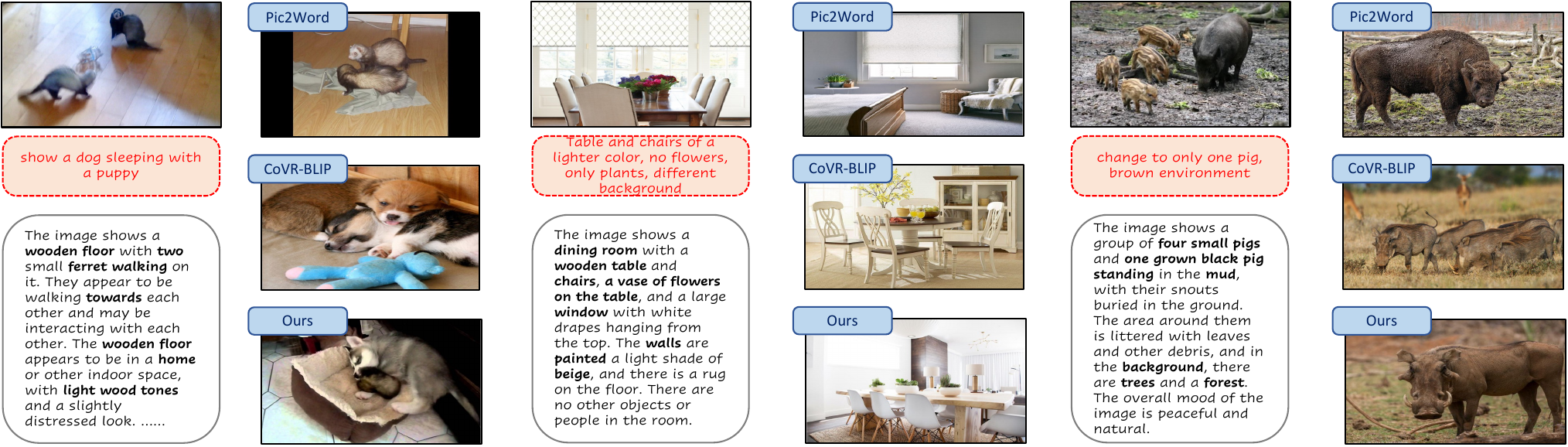}
   \caption{
   Qualitative Comparison between Pic2Word~\cite{saito2023pic2word}(top row), CoVR-BLIP~\cite{ventura2023covr} (mid-row) and our proposed method (bottom-row) in zero-shot CoIR task. Here in all three examples from CIRR test set, we observe that using only reference image and change text (in red) Pic2Word~\cite{saito2023pic2word} and CoVR-BLIP~\cite{ventura2023covr} struggle to correctly retrieved target video (top and mid row). In comparison, our approach leveraging detailed descriptions is accurately retrieving the target video with most relevant contextual match with reference video (bottom row). Best viewed zoomed in.
   }
   \label{fig:coir_qualitative_figure}
\end{figure}

\subsection{Composed Image Retrieval}
We present state-of-the-art comparison in \cref{tab:cirr_results} and \cref{tab:fashioniq_results} on CIRR~\cite{cirr} and FashionIQ~\cite{fashioniq}, respectively. Here, the target embeddings are w.r.t a single image. 
We report results in two settings: zero-shot and transfer learning on respective datasets. In the zero-shot setting, we use our model trained on WebVid-CoVR and directly apply it to these respective benchmarks. In both cases, our approach achieves superior performance, indicating that the visual description of input image improves the alignment capability of the model between input and target examples. ~\cref{fig:coir_qualitative_figure} presents a qualitative comparison with existing works on example samples from the CIRR~\cite{cirr} test set.

\section{Conclusion}
In conclusion, our proposed method effectively contributes to the Composed Video Retrieval (CoVR) and Composed Image Retrieval (CoIR) tasks by integrating detailed visual descriptions. The descriptions are generated from advanced vision-language conversational models with relative change text and visual features that help our approach successfully bridge a critical gap in the retrieval process. The enhanced contextual understanding and richer content interpretation offered by our approach proves to be pivotal in outperforming existing state-of-the-art models by a considerable margin of 7\%. Furthermore, the robustness of our method is demonstrated by its strong performance in zero-shot setups, particularly in composed image retrieval (CoIR) tasks.

\section{Acknowledgement}
The computations were enabled by resources provided by the National Academic Infrastructure for Supercomputing in Sweden (NAISS) at Alvis partially funded by the Swedish Research Council through grant agreement no. 2022-06725, the LUMI supercomputer hosted by CSC (Finland) and the LUMI consortium, and by the Berzelius resource provided by the Knut and Alice Wallenberg Foundation at the National Supercomputer Centre.

{
    \small
    \bibliographystyle{ieeenat_fullname}
    \bibliography{arxiv}
}

\end{document}